\pdfoutput=1

\documentclass[11pt]{article}

\usepackage[]{acl}

\usepackage{times}
\usepackage{latexsym}

\usepackage[T1]{fontenc}

\usepackage[utf8]{inputenc}

\usepackage{color}
\usepackage{booktabs}
\usepackage{multirow}

\usepackage{graphicx} 
\usepackage{float}
\usepackage{subfigure} 
\usepackage{amssymb}

\usepackage{hyperref}
\usepackage{xspace}

\usepackage{float}
\usepackage{amsmath}
\usepackage{bm}

\usepackage{amssymb}
\usepackage{pifont}
\newcommand{\cmark}{\ding{51}}%
\newcommand{\xmark}{\ding{55}}%
\usepackage{enumitem}

\usepackage{tcolorbox}
\usepackage{xcolor}
\usepackage{colortbl}
\usepackage{fontawesome5}
\usepackage{caption}
\usepackage{microtype}
\usepackage{arydshln}
\usepackage{afterpage}

\definecolor{backred}{RGB}{255, 190, 190}
\definecolor{backblue}{RGB}{210, 230, 250}
\usepackage{microtype}
\usepackage{fontawesome5}

\newcommand{\ours}{\textsc{FinDVer}\xspace}
\newcommand{\ieset}{\textsc{FDV-IE}\xspace}
\newcommand{\mathset}{\textsc{FDV-Math}\xspace}
\newcommand{\knowset}{\textsc{FDV-Know}\xspace}

\newcommand{\modelnum}{16\xspace}
\newcommand{\orgnum}{9\xspace}
\newcommand{\nexample}{2,400\xspace}
\newcommand{\best}{76.2\%\xspace}
\newcommand{\human}{93.3\%\xspace}
\newcommand{\bestmodel}{GPT-4o\xspace}

\newcommand{\minicheck}{\textsc{LLM-AggreFact}\xspace}
\newcommand{\testmini}{\emph{testmini}\xspace}
\newcommand{\test}{\emph{test}\xspace}

\newcommand{\eg}{\hbox{\emph{e.g.,}}\xspace}
\newcommand{\ie}{\hbox{\emph{i.e.,}}\xspace}
\newcommand{\entail}{\emph{``entailed''}\xspace}
\newcommand{\refute}{\emph{``refuted''}\xspace}

\newcommand{\down}[1]{\textcolor{blue}{(-#1)}}

\title{\ours: Explainable Claim Verification over Long and\\ Hybrid-Content Financial Documents}

\author{Yilun Zhao \quad Yitao Long \quad Yuru Jiang \quad Chengye Wang \quad Weiyuan Chen \\\bf{Hongjun Liu \quad Yiming Zhang \quad  Xiangru Tang \quad Chen Zhao \quad Arman Cohan} \vspace{8pt}\\
Yale NLP - \ours Team\\
}

\begin{document}
\maketitle
\begin{abstract}
We introduce \ours, a comprehensive benchmark specifically designed to evaluate the explainable claim verification capabilities of LLMs in the context of understanding and analyzing long, hybrid-content financial documents. 
\ours 
contains \nexample expert-annotated examples, 
divided into three subsets: information extraction, numerical reasoning, and knowledge-intensive reasoning—each addressing common scenarios encountered in real-world financial contexts.
We assess a broad spectrum of LLMs under long-context and RAG settings. 
Our results show that even the current best-performing system, \bestmodel, still lags behind human experts. 
We further provide in-depth analysis on long-context and RAG setting, Chain-of-Thought reasoning, and model reasoning errors, offering insights to drive future advancements.
We believe that \ours can serve as a valuable benchmark for evaluating LLMs in claim verification over complex, expert-domain documents.

\centering
{
 \href{https://github.com/yilunzhao/FinDVer}{{\faGithub{}}\texttt{\small ~github.com/yilunzhao/FinDVer}}
}
\vspace{4pt}
\end{abstract}

\section{Introduction}
In today's information explosion era, the responsibility of verifying the truthfulness of the item is often passed on to the audience.unverified claims about a company's financial performance frequently circulate in online media, potentially misleading investors. Therefore, it is crucial to verify these claims using the companies' original financial documents (\ie earnings reports and regulatory filings). 
Recent advancements in Large Language Models (LLMs) have attracted significant attention due to their capabilities in solving a broad range of tasks~\cite{Touvron2023Llama2O, Jiang2023Mistral7, OpenAI2023GPT4TR}. 
However, it remains particularly difficult for
applying them to document-grounded claim verification in real-world financial domains due to the following two reasons:

\begin{figure}[!t]
    \centering
    \includegraphics[width = \linewidth]{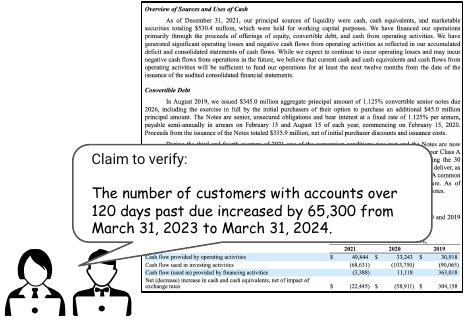}
    \caption{An example from the \emph{numerical reasoning} subset of the \ours benchmark. To verify the claim, the LLM is required to first locate claim-relevant data points within long and hybrid-content financial documents, and then apply numerical reasoning over the extracted data points for claim verification.
    }
    \label{fig:example}
\end{figure}
\begin{table*}[!t]
\centering
\renewcommand{\arraystretch}{1.05}
\resizebox{\textwidth}{!}{%
\addtolength{\tabcolsep}{-0.2em}
\begin{tabular}{lllc>{\centering\arraybackslash}p{1.5cm}>{\centering\arraybackslash}p{1.5cm}}
\toprule
\multirow{2}{*}{Dataset}     & \multirow{2}{*}{Input Context} & \multirow{2}{*}{Annotation / Data Creation} & \multirow{2}{*}{\# Label} & $w.$ Explanation? & Reason-Intensive?\\ 
\midrule
PubHealthTab~\cite{akhtar-etal-2022-pubhealthtab} & Wikipedia table & Crowd-sourced & 4 & \xmark & \xmark \\
\cdashline{5-6}

\textsc{TabFact}~\cite{Chen2020TabFact} & Wikipedia table & Crowd-sourced & 2 & \xmark & \cmark \\ 
\textsc{InfoTabS}~\cite{gupta-etal-2020-infotabs} & Wikipedia table & Crowd-sourced & 3 & \xmark & \cmark \\ 
\textsc{SciTab}~\cite{lu-etal-2023-scitab} & Scientific table & Expert \& InstructGPT & 3 & \xmark & \cmark \\ 
\noalign{\vskip 0.5ex}\hdashline\noalign{\vskip 0.5ex}
\textsc{HoVer}~\cite{jiang-etal-2020-hover} & Wikipedia articles & Crowd-sourced & 2 & \xmark & \xmark \\
\textsc{DocNLI}~\cite{yin-etal-2021-docnli} & News article & From summrization datasets & 2 & \xmark & \xmark \\ 
ContractNLI~\cite{koreeda-2021-contractnli-dataset} & Contract & Expert \& Crowd-sourced & 2 & \xmark & \xmark \\ 
\minicheck~\cite{tang2024minicheck} & Doc from various domains & From existing benchmarks & 2 & \xmark & \xmark\\
\textsc{WiCE}~\cite{kamoi-etal-2023-wice} & Wikipedia article & Crowd-sourced & 3 & \xmark & \xmark \\ 
\textsc{AmbiFC}~\cite{glockner-etal-2024-ambifc} & Wikipedia article & Crowd-sourced & 3 & \xmark & \xmark\\
\textsc{ClaimDecomp}~\cite{chen-etal-2022-generating} & Political article & Expert & 6 & \xmark & \xmark \\

\textsc{SciFact}~\cite{wadden-etal-2020-fact} & Scientific paper abstracts & Expert & 2 & \xmark & \cmark \\ 

\cdashline{5-6}

LIAR++~\cite{russo-etal-2023-benchmarking} & Political article & From fact-check website & 2 & \cmark & \xmark \\
FullFact~\cite{russo-etal-2023-benchmarking} & Web page & From fact-check website & 2 & \cmark & \xmark \\
\textsc{PubHealth}~\cite{akhtar-etal-2022-pubhealthtab} & Health web page & From fact check website & 4 & \cmark & \xmark\\
\midrule
\textbf{\ours (ours)} & Long financial doc with tables & Expert & 2 & \cmark & \cmark \\ 
\bottomrule
\end{tabular}
}
\caption{Comparison between \ours and existing table- or document-grounded claim verification benchmarks. 
\ours is distinguished by four unique characteristics: 
(1) \emph{Expert Annotation}: It is annotated by financial experts to ensure high data quality;
(2) \emph{Complex Document Comprehension}: It requires interpreting a mix of textual and tabular data within a long-context financial document;
(3) \emph{Examination on Reasoning-Process Explanation}: It enhances claim verifications with detailed explanations about the reasoning process, increasing the benchmark's practical value;
and (4) \emph{Diverse Reasoning for Real-world Scenarios}: It incorporates various reasoning challenges, such as extracting complicated information, performing numerical calculations, and applying professional knowledge.
}
\label{tab:dataset_comparison}
\end{table*}
First, financial documents are typically long, intricate and dense, and they include both quantitative tables and qualitative text~\cite{chen-etal-2021-finqa, zhu-etal-2021-tat, zhao-etal-2022-multihiertt, li-etal-2022-finmath, zhao-etal-2023-robut, koncelkedziorski2024bizbench}. 
Extracting and analyzing claim-relevant data from these documents requires complicated document comprehension abilities and professional knowledge in financial domains. Moreover, the type of reasoning involved encompasses various unique aspects that are less studied, necessitating a dedicated approach to evaluation and application.

Second, in the financial domain, where decisions often involve significant stakes, it is often critical to provide clear and comprehensible rationales for any claim verification decisions~\cite{atanasova-etal-2020-generating-fact, atanasova-etal-2023-faithfulness}. 
However, existing \emph{context-grounded} claim verification benchmarks~\cite{Chen2020TabFact, kamoi-etal-2023-wice, lu-etal-2023-scitab, glockner-etal-2024-ambifc} primarily focus on the task of entailment classification and do not evaluate the reasoning process. This hinders the practical application and evaluation of LLMs in real-world scenarios. 

In response to the aforementioned pressing need, we present \textbf{\ours}, 
a comprehensive and domain expert-annotated explainable claim verification benchmark that first explores in the context of financial documents. 
The LLMs are tasked with generating explanations of their reasoning to verify claims labeled as entailed or refuted and, based on the information in the provided document, which contains both textual and tabular data.
To identify the common reasoning-intensive scenarios in claim verification based on financial documents, we engage with domain experts and conducted a preliminary study. This helped us determine three key types of scenarios that frequently arise in real-world settings:
\emph{information extraction}, \emph{numerical reasoning}, and \emph{knowledge-intensive reasoning}. 
For each scenario, we construct an evaluation set.
Each example in our dataset is annotated with detailed supporting evidence and step-by-step reasoning-process explanations.

We evaluate a wide spectrum of open- and closed-source LLMs, specifically, \modelnum models from \orgnum organizations. The documents in our benchmark are exceedingly long; therefore, we employ two widely adopted real-world application settings—\emph{retrieval augmented generation} (RAG) and \emph{long-context}—in this study.
The experimental results indicate that even the existing best-performing LLM (\ie \bestmodel) still significantly lags behind human experts (\best versus \human), demonstrating the challenges of our proposed benchmark.
Our contributions are summarized below:
\begin{itemize} [leftmargin=*]
\itemsep0em 
\item We introduce \ours, the first comprehensive context-grounded claim verification benchmark for financial domains, presenting new challenges for state-of-the-art LLMs.
\item We conduct an extensive evaluation that encompasses a wide range of LLMs, comprehensively assessing the capabilities and limitations of existing LLMs in our task.
\item We provide in-depth analysis on Long-context setting, RAG setting, Chain-of-Thought reasoning, and model reasoning errors, offering valuable insights to drive future advancements.

\end{itemize}

\begin{figure*}[!t]
\centering
\includegraphics[width=\textwidth]{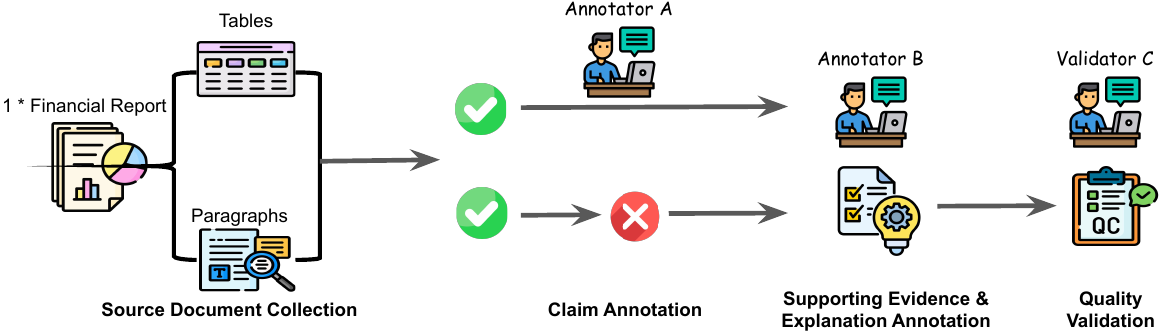}
\caption{An overview of the \ours construction pipeline. 
We collect and process quarterly and annual reports from companies, which contain both tables and text, as source financial documents (\S\ref{sec:document-collection}). 
For each financial document, expert annotators are first tasked with annotating the \entail claims. Next, they are asked to perturb these \entail claims to introduce factual errors, making the original claims into \refute claims for the purpose of \refute claim annotation (\S\ref{sec:claim-annotaion}).
For each claim, the annotators are required to provide supporting evidence and an explanation of their reasoning process (\S\ref{sec:reasoning-annotation}).
Finally, each annotated example undergoes quality validation by a separate expert annotator (\S\ref{sec:quality-validation}). This designed data construction pipeline ensures the high quality of \ours.
}
\label{fig:pipeline}
\end{figure*}
\section{Related Work}
\paragraph{Claim Verification Benchmark}
Claim verification is a well-established research area with two main settings. The first is the open-domain setting, which involves using an external retriever to find the most relevant information from a large corpus to verify claims~\cite{vlachos-riedel-2014-fact, thorne-etal-2018-fever, feverous, wadden-etal-2022-scifact, rangapur2024finfact, ma2024exfever}. The second setting is context-grounded claim verification, which requires models to verify claims based on the provided document context~\cite{Chen2020TabFact, kamoi-etal-2023-wice, lu-etal-2023-scitab, glockner-etal-2024-ambifc}.
This work focuses on the second setting, as it allows us to eliminate variability and dependency on the retriever's performance, thereby focusing on the evaluation of LLM performance on on accurately verifying claims within a given context.
However, as illustrated in \autoref{tab:dataset_comparison}, existing context-grounded claim verification benchmarks have three notable limitations: they typically 1) focus on general domains, overlooking the specific challenges and intricacies present in specialized fields, 2) focus solely on entailment classification and do not evaluate the reasoning processes of models, 3) do not involve claims that require intensive reasoning and complicated document comprehension. These limitations hinder their effectiveness for evaluating LLMs in real-world practice.

\paragraph{Financial Evaluation Benchmark}
NLP techniques have been applied to various financial tasks~\cite{xie2023pixiu, xie2024finbenholisticfinancialbenchmark}, such as
named entity recognition~\cite{salinas-alvarado-etal-2015-domain, shah2023finer}, 
sentiment analysis~\cite{malo2013good, 10.1145/3184558.3192301},
stock movement prediction~\cite{xu-cohen-2018-stock, 10.1145/3269206.3269290,10020720, xie2023wallstreetneophytezeroshot}, 
and summarization~\cite{zhou-etal-2021-trade, mukherjee-etal-2022-ectsum, liu-etal-2022-long}. 
More recently, there has been an increasing focus on tasks involving financial documents (\eg annual reports and regulatory filings), which are crucial for providing insights into a company's performance and strategies. 
Several QA benchmarks have been proposed to evaluate models' performance in answering questions based on financial documents, with a particular focus on numerical reasoning~\cite{chen-etal-2021-finqa, zhu-etal-2021-tat, zhao-etal-2022-multihiertt, chen-etal-2022-convfinqa, koncelkedziorski2024bizbench, zhao-etal-2024-knowledgefmath, zhao-etal-2024-docmath}.
Despite these advancements, there remains a significant gap in the exploration of claim verification tasks within the financial domain.
While the recent \textsc{Fin-Fact} benchmark~\cite{rangapur2024finfact} addresses explainable multimodal financial fact-checking, it primarily focuses on open-domain scenarios. Verifying claims derived from financial documents is crucial, as inaccuracies can significantly influence investment decisions and market perceptions. To bridge this gap, we introduce \ours, the first context-grounded claim verification benchmark, specifically designed for real-world financial document comprehension.

\section{\ours Benchmark}\label{sec:data}
\ours provides a robust evaluation benchmark for reasoning-intensive and explainable claim verification over long and hybrid-content 
financial documents. 
\autoref{fig:detailed-example} in appendix presents a detailed dataset example of our benchmark.
In the following subsections, we present an overview of the \ours construction pipeline in \autoref{fig:pipeline}; and detail the task formulation, data construction, and quality validation process.

\subsection{Task Formulation}
We formally define the task of \ours within the context of LLMs as follows: Consider a \emph{single} financial document $d$, containing textual data $P$ and tabular data $T$, associated with a claim $c$ that needs verification.
The expert-annotated data we collect supports the following two tasks:

\paragraph{Entailment Classification}
The LLM is required to determine the entailment label $\ell \in \mathcal{L} = \{\text{\entail}, \text{\refute}\}$, based on the given hybrid-content financial document:

\begin{equation}
\label{eq:labeling}
    \ell = \arg\max_{\ell \in \mathcal{L}} P_{~\mathbf{LLM}}(\ell~|~P, T, c)
\end{equation}

\paragraph{Reasoning-process Explanation Generation}
The model is required to generate a natural language explanation $e$: 

\begin{equation}
\label{eq:formulation}
    e = \arg\max_e P_{~\mathbf{LLM}}(e~|~P, T, c)
\end{equation}

\noindent which articulates the reasoning process behind the validity of the provided claim $c$, based solely on the provided textual content $P$ and tabular content $T$ within the financial document. \vspace{8pt}

Notably, some claim verification systems,  
particularly those developed prior to the era of LLMs and for previous datasets that did not require explanation generation~\cite{Chen2020TabFact, yin-etal-2021-docnli, koreeda-2021-contractnli-dataset}, 
might not explicitly perform explanation generation. 
Instead, they directly output the final label. For such systems, \ours can also be used for evaluation by focusing on the entailment classification task.

\subsection{\ours Subset Design}
\ours is designed to mirror the real-world challenges encountered in the financial domain. Therefore, we ensure that the included annotators are financial experts with professional experience in comprehending and processing financial documents. \autoref{tab:candidate_profiles} in appendix presents the detailed annotator biographies for \ours annotation. 

To identify the common reasoning-intensive scenarios in claim verification based on financial documents, we engaged with domain experts and conducted a preliminary study. This helped us determine three key types of scenarios that frequently arise in real-world settings.
Accordingly, we have created three corresponding subsets of \ours. 

\noindent (1) \textbf{\ieset} (\emph{information extraction}), which involves extracting information from both \emph{textual} and \emph{tabular} content within a \emph{long-context} document. \vspace{2pt}

\noindent (2) \textbf{\mathset} (\emph{numerical reasoning}), which necessitates performing \emph{calculations} or \emph{statistical analysis} based on data within the document. \vspace{2pt}

\noindent (3) \textbf{\knowset} (\emph{knowledge-intensive reasoning}), which requires integrating \emph{external domain-specific knowledge} or \emph{regulations} for claim verification. \vspace{2pt}

\subsection{Source Document Collection}\label{sec:document-collection}
Similar to \citet{zhao2023docmatheval}, we use the quarterly (Form 10-Q) and annual reports (Form 10-K) of companies as the source documents, which are publicly available in the open-source database\footnote{\url{https://www.sec.gov/edgar/search/}} of the U.S. Securities and Exchange Commission. 
We collect a total of 523 documents that were first released between January 1 to April 30, 2024, which is after the cutoff date of most pretraining corpora for training foundation models. This helps to alleviate issues related to data memorization to some extent.
After collecting the raw HTML-format documents, we utilize the SEC API\footnote{\url{https://sec-api.io/}}, a commercial platform API for extracting financial document content, to  process the collected documents, obtaining documents with both textual and tabular data. 

\begin{table*}[!t]
\centering 
\small
\renewcommand{\arraystretch}{1.1}
\newcolumntype{L}{>{\raggedright\arraybackslash}p{2.5cm}}
\newcolumntype{R}{>{\raggedleft\arraybackslash}p{2.5cm}}
\newcolumntype{C}{>{\centering\arraybackslash}p{2.5cm}}
\resizebox{\textwidth}{!}{
\begin{tabular}{lRRRR}
\toprule
\textbf{Property}  & \multicolumn{1}{C}{\bf \ieset} & \multicolumn{1}{C}{\bf\mathset} & \multicolumn{1}{C}{\bf\knowset}\\
\midrule
Real-world scenarios in financial domains & \multicolumn{1}{C}{information extraction} & \multicolumn{1}{C}{numerical reasoning} & \multicolumn{1}{C}{knowledge-intensive reasoning} \\
\midrule
$\#$ Document & 322 & 300 & 314
\\
\quad Doc Length \texttt{(Median/Avg/Max)} & 39K / 40K / 69K & 38K / 40K / 68K & 38K / 40K / 70K \\
$\#$ Tables per document  \texttt{(Median/Avg)}  &  41.5 / 42.6 & 38.0 / 40.7 &  41.0 / 41.9
\\
\noalign{\vskip 0.5ex}\hdashline\noalign{\vskip 0.5ex}
Claim length  \texttt{(Median/Avg)}  & 45.0 / 45.9 & 24.0 / 24.5 & 46.0 / 46.4 \\
$\#$ Text evidence per claim \texttt{(Avg)} & 2.2 & 1.0 & 2.9  \\
$\#$ Table evidence per claim \texttt{(Avg)} & 0.7 &  0.9 & 0.8 \\

Explanation length  \texttt{(Median/Avg)} & 70.0 / 73.1 & 74.0 / 76.2 & 96.0 / 100.7\\

\midrule
Benchmark size ($\#$ Claims)\\
\quad \testmini size  & 250 & 250 & 200\\
\quad \test size  & 600 & 600 & 500\\
\toprule
\end{tabular}
}

\caption{Basic statistics of the \ours benchmark.}
\label{tab:data_statistics}
\end{table*}
\subsection{Claim Annotation} \label{sec:claim-annotaion}
\paragraph{Entailed Claim Annotation}
To address the potential bias concerning the position of evidence within the documents, we initiate the process by randomly sampling multiple document contexts from the given document. Annotators are then tasked with creating \entail claims based on the textual and tabular data within these contexts. 
The annotators are instructed to simulate real-world document comprehension scenarios, ensuring the annotated claims are representative of practical financial analysis and align with the scenarios defined by the corresponding subsets.
Annotators are then tasked with carefully locating all evidence (\ie indices of relevant paragraphs and tables) within the entire document that support the claims, which are used for the subsequent data validation.

\paragraph{Refuted Claim Annotation}
Following established practices in the field~\cite{wadden-etal-2020-fact, Chen2020TabFact, lu-etal-2023-scitab}, and since directly obtaining \refute types is difficult, we instead perturb the original \entail claims into \refute claim through expert annotation.
Specifically, expert annotators first create an \entail claim using the same procedure detailed in the ``Entailed Claim Annotation'' paragraph.
The annotators are then instructed to perturb the \entail claim to introduce factual errors that are directly contradicted by the annotated evidence, and rewrite the annotated reasoning-process explanation.

\subsection{Explanation Annotation} \label{sec:reasoning-annotation}
After finishing the claim annotation, we pass it to another annotator for explanation annotation. The annotators are required to first read the claim carefully and annotate a detailed explanation of the reasoning process. Such reasoning-process explanations allow for a granular and informative evaluation of model outputs, helping future work identify reasoning errors and provide more accurate error feedback. 
We compare the entailment label annotated in this step with those in the claim annotation step. A third annotator is introduced if the two annotation versions are different. In practice, we achieve an inter-annotator agreement of 90.3\% for entailment label annotation.

During our pilot annotation phase, we observed variability in the format of reasoning-process explanation annotated by different annotators, which made the dataset less standardized. To ensure consistency and clarity in our benchmark, we developed a predefined template for annotators to follow. 
Specifically, annotators are required to commence with the \textbf{extraction of relevant information} phase, where they need to list all claim-relevant information in a numbered list. Subsequently, they are required to annotate the \textbf{reasoning over the extracted information} segment in a step-by-step manner. For each step, they should elucidate the associated reasoning. Finally, they annotate the \textbf{entailment label} feature.

\subsection{Data Quality Validation}\label{sec:quality-validation}
To ensure the high quality of our annotated data, for every annotated example, we first use the GPT-4o model to proofread and refine both the annotated statement and explanation. Then a qualified annotator is assigned to validate several key aspects:
(1) the claim and reasoning-process explanation should be grammatically correct and free of spelling errors;
(2) the claim should be closely related to financial domains and meaningful in real-world scenarios;
(3) the annotated evidence should be relevant to the claim and complete enough to verify it;
(4) the entailment label of the claim should be supported by the annotated evidence;
and (5) the reasoning-process explanation should correctly interpret the extracted evidence and apply appropriate reasoning steps to correctly
verify the claim.
The validators are asked to revise examples that do not meet these standards. In practice, 347 out of \nexample initial examples were revised by the validators.
We also report the human evaluation scores over 100 sampled examples. As illustrated in \autoref{tab:annotation_aggrement} in appendix,
\ours has a high annotation quality.

\subsection{Dataset Preparation and Release}
\autoref{tab:data_statistics} presents an overview of the primary statistics for our dataset. 
\ours is divided into two subsets: \testmini and \test. 
The \testmini subset is designed specifically for model development and validation, while the \test subset is reserved for formal evaluation. 
To mitigate the risk of data contamination~\cite{jacovi-etal-2023-stop, deng-etal-2024-unveiling}, we do not release ground-truth-related annotation features for the \test set publicly. Instead, we will develop and maintain an online evaluation platform where researchers can test their models and participate in a public leaderboard.
\section{Experiment Setup}
We next present the experimental setup, covering the evaluated LLMs, long-context and RAG setups, implementation details, and the measurement of human-level performance.
\subsection{Experimented LLMs}
We examine the performance of LLMs across two distinct categories on \ours: 
(1) \textbf{Proprietary LLMs}, including 
GPT-4o~\cite{openai2024gpt4o}, 
Gemini-1.5-Pro~\cite{geminiteam2024gemini}, 
and Claude-3.5-Sonnet~\cite{claude3}; 
and (2) \textbf{Open-source LLMs}, including 
Llama-3.1\&3.2~\cite{meta2024introducing},  
Qwen-2\&2.5~\cite{qwen2, qwen2.5}, 
Mistral \& Mixtral \& Mathstral \& Ministral~\cite{jiang2023mistral, jiang2024mixtral},
InternLM2.5~\cite{cai2024internlm2}, 
DeepSeek-V2-Lite~\cite{deepseekai2024deepseekv2},
and GLM~\cite{du-etal-2022-glm}.
\autoref{tab:model} in Appendix presents the details of evaluated models (\ie organizations, release time, max context length, and model version). 
The experiments with open-source LLMs use the vLLM framework~\cite{kwon2023efficient}. For all experiments, the temperature is set to 1.0, and the maximum output length is 512 tokens.
We employ Chain-of-Thought (CoT) prompting methods~\cite{wei2022chain} for the main experiments. In this approach, the model first generates a detailed reasoning process to verify each claim and then provides an entailment label based on this reasoning. The specific prompts used for this CoT methodology are shown in \autoref{fig:cot_prompt} in appendix.

\begin{table*}[!t]
\centering
\resizebox{0.9\textwidth}{!}{%
\begin{tabular}{lrcccccccccccccccc}
\toprule
\multirow{2}{*}{\textbf{Model}} & \multirow{2}{*}{\textbf{Size}} & \multicolumn{2}{c}{\textbf{\ieset}} & \multicolumn{2}{c}{\textbf{\mathset}} & \multicolumn{2}{c}{\textbf{\knowset}} & \multicolumn{2}{c}{\textbf{Average}} \\
\cmidrule(lr){3-4} \cmidrule(lr){5-6} \cmidrule(lr){7-8} \cmidrule(lr){9-10}
& & Testmini & Test & Testmini & Test & Testmini & Test & Testmini & \textbf{Test} \\
\midrule
Human Non-Expert & & \multicolumn{2}{c}{90.0} & \multicolumn{2}{c}{85.0} & \multicolumn{2}{c}{85.0} & \multicolumn{2}{c}{86.7}\\
Human Expert & & \multicolumn{2}{c}{95.0} & \multicolumn{2}{c}{90.0} & \multicolumn{2}{c}{95.0} & \multicolumn{2}{c}{93.3}\\
\midrule

DeepSeek-V2-Lite & 16B & 60.4 & 57.7 & 64.0 & 58.5 & 56.0 & 58.8 & 60.1 & 58.3 \\
Llama-3.2 & 3B &  65.6 & 60.3 & 55.6 & 57.0 &53.0 & 57.8 & 58.4 & 58.4 \\
Mathstral & 7B &  66.4 & 61.2 & 58.0 & 59.8 &59.0 & 62.0 & 61.3 & 60.9 \\
Mistral-v0.3 & 7B &  71.2 & 67.3 & 61.2 & 59.3 &72.5 & 65.6 & 68.0 & 64.0 \\
InternLM2.5 & 7B &  71.2 & 70.2 & 61.6 & 56.8 &65.0 & 66.2 & 66.0 & 64.3 \\
Llama-3.1 & 8B &  70.8 & 72.2 & 64.0 & 58.3 &64.0 & 64.2 & 66.4 & 64.9 \\
Qwen2 & 7B &  72.8 & 69.3 & 63.6 & 61.2 &65.5 & 69.4 & 67.4 & 66.5 \\
Ministral & 8B &  74.4 & 70.5 & 64.8 & 62.8 &63.5 & 66.8 & 67.8 & 66.7 \\
GLM-4 & 9B &  75.6 & 72.3 & 68.0 & 64.2 &70.5 & 70.6 & 71.4 & 68.9 \\
Qwen2.5 & 7B &  77.2 & 71.5 & 68.4 & 68.2 &71.5 & 71.2 & 72.4 & 70.2 \\
Claude-3.5-Sonnet & -- &  74.8 & 76.3 & \cellcolor{blue!25}{77.2} & 69.5 &66.0 & 64.4 & 73.1 & 70.4 \\
Gemini-1.5-Pro & -- &  75.2 & 77.5 & 69.6 & 70.8 &69.0 & 70.8 & 71.4 & 73.2 \\
Llama-3.1 & 70B &  \cellcolor{blue!15}{78.8} & \cellcolor{blue!15}{78.8} & 66.8 & 66.2 &\cellcolor{blue!25}{80.5} & \cellcolor{blue!25}{79.2} & \cellcolor{blue!5}{75.0} & 74.5 \\
Qwen2.5 & 72B &  \cellcolor{blue!25}{80.8} & 77.2 & \cellcolor{blue!5}{72.8} & \cellcolor{blue!5}{71.0} &73.0 & \cellcolor{blue!15}{77.0} & \cellcolor{blue!25}{75.7} & \cellcolor{blue!5}{74.9} \\
Mistral-Large & 123B &  \cellcolor{blue!5}{78.4} & \cellcolor{blue!5}{78.3} & 72.4 & \cellcolor{blue!25}{73.5} &\cellcolor{blue!15}{73.5} & 75.6 & 74.8 & \cellcolor{blue!15}{75.8} \\
GPT-4o & -- &  78.0 & \cellcolor{blue!25}{80.7} & \cellcolor{blue!15}{74.0} & \cellcolor{blue!15}{71.3} &\cellcolor{blue!5}{73.5} & \cellcolor{blue!5}{76.8} & \cellcolor{blue!15}{75.3} & \cellcolor{blue!25}{76.2} \\

\bottomrule

\end{tabular}
}
\caption{
Accuracy of entailment classification on the \ours \testmini and \test splits. Results are reported for LLMs using \emph{CoT} prompting under the \emph{RAG} setting. The average accuracy on the test set is used as ranking indicator.
}
\label{tab:results}
\end{table*}

\subsection{Long-Context and RAG Settings}
As presented in \autoref{tab:data_statistics}, the documents within our benchmark are notably lengthy. To effectively handle this, we have implemented two real-world application settings that are widely recognized for their utility in dealing with extensive texts.
For \textbf{Long-context Setting}, we input the entire financial document into the model. We limit our evaluation to those models that have a context window larger than 100,000 tokens, which exeeds the maximum length of the included financial document.
For \textbf{RAG Setting}, we leverage the current best-performing embedding models (\ie OpenAI's \texttt{text-embedding-3-large}) to retrieve the top-$10$ paragraphs or tables that are most relevant to the claims. These elements are then concatenated in their original order as found in the document before being fed into the model.

\subsection{Implementation Details} \label{appendix:implementation-details}

\paragraph{Input Tabular Data Serialization}
Building on previous research that assessed LLMs on tasks involving tabular data~\cite{chen-2023-large, zhao-etal-2023-qtsumm, zhao-etal-2023-investigating}, we introduce our methodology for processing tables within documents. Our approach encodes table structures by delineating column headers and cell values with vertical bars (|) and separating rows with line breaks. 
This flattened format enables the direct input of tabular data into LLMs. Initial experiments with models such as Qwen-2.5 and Llama-3.1 demonstrate their ability to effectively interpret this simplified table encoding. Nonetheless, we encourage future research to explore more advanced encoding techniques for tabular data to further improve LLM comprehension and performance on complex tables.

\paragraph{Model Response Processing} 
Following previous work~\cite{lu2024mathvista}, we adopt LLM for processing model response.
Specifically, we utilize GPT-4o-mini to extract labels from the LLM output, which can be either \entail, \refute or \emph{``none''}. The \emph{``none''} label typically indicates that the LLM output contains nonsensical symbols or unintelligible text rather than meaningful content. In cases where the output is labeled as \emph{``none''}, we assign the final label by making a random guess.

\subsection{Human-level Performance Measurement}
To provide a rough but informative estimate of human-level performance by non-experts and experts on \ours, we randomly sampled 5 documents $\times $ 4 claims / document = 20 claims  from each validation subset, totaling 60 claims. We enroll two experts (\ie professionals with CFA license) and two non-experts (\ie undergraduate students majored in computer science) to individually verify the claims by providing the NL explanations. \autoref{tab:results} presents the human-level performance.

\section{Experiment Results} \label{sec:rq1}
This section discusses the experiment results on \ours, including our main findings, ablation studies, and error analysis.

\subsection{Main Findings}\label{sec:rq1}
\autoref{tab:results} display the results for \ours under the RAG setting. 
We observe a significant accuracy gap between human experts and the evaluated LLMs. Notably, \bestmodel, the best-performing LLM, achieves an average accuracy of only \best, which is markedly lower than the \human accuracy achieved by human experts. This gap underscores the complexity and challenges presented by \ours.
For open-source LLMs, frontier models like Llama-3.1, Qwen2.5, and Mistral have achieved performance levels on par with proprietary counterparts. For example, the Llama-3.1-70B model surpasses GPT-4o on the \knowset dataset. This progress underscores the promising potential of open-source models for reasoning-intensive tasks within specialized domains.

\subsection{Long-Context Setting Analysis}
\begin{figure}[!t]
    \centering
    \includegraphics[width = \linewidth]{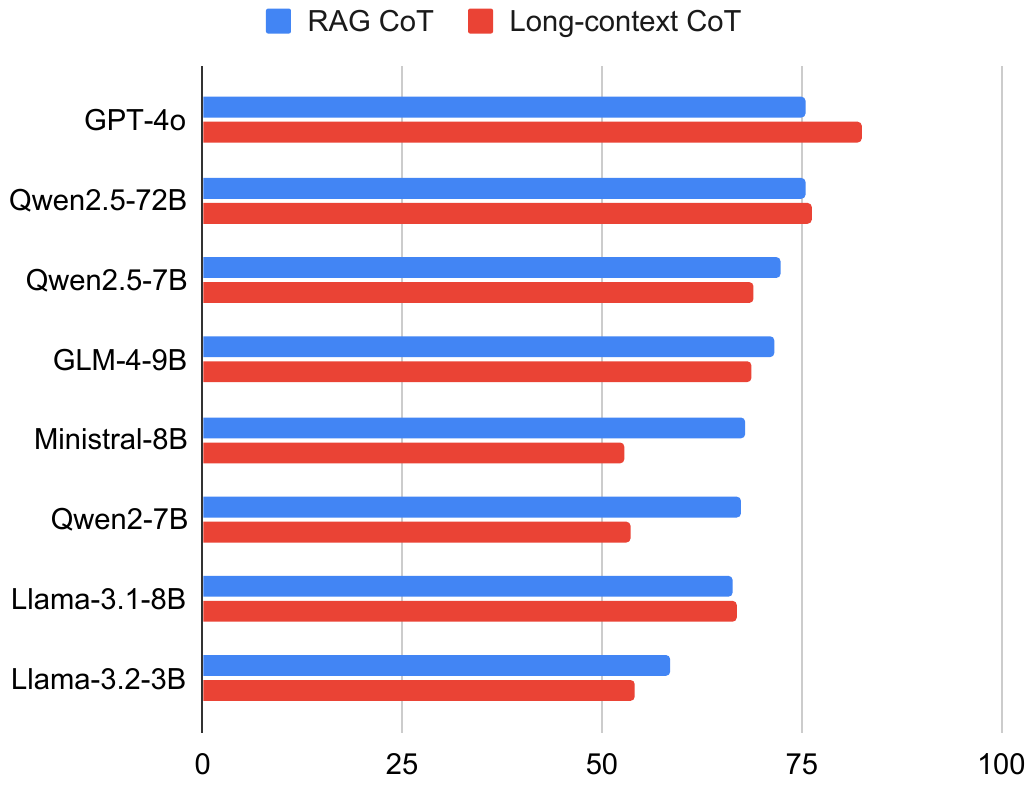}
    \caption{Comparison of LLM performance on the \testmini split in long-context versus RAG settings using the CoT prompting method.
    }
    \label{fig:rag_lc}
\end{figure}

We next provide an analysis of model performance in the long-context setting.
\autoref{fig:rag_lc} presents a comparison of model performance in long-context versus RAG settings.
Frontier LLMs, such as GPT-4o and Qwen2.5-72B, perform better in the long-context setting, highlighting their capability to effectively handle extended input lengths.
An encouraging trend is also observed in the progress of models adapted for long-context settings. For instance, the earlier Qwen2-7B model shows a pronounced performance gap between the long-context and RAG settings; while the latest Qwen2.5-7B model significantly narrows this performance gap. This suggests that recent developments in model architecture and training techniques are enhancing models' performance in long-context scenarios.



\subsection{RAG Setting Analysis}
This subsection explores the impact of evidence retrieval accuracy on the overall performance of LLMs. We assess LLM performance using three different retrieval methods, \ie BM25~\cite{robertson1995okapi}, Contriever~\cite{izacard2022unsupervised}, and OpenAI's \texttt{text-embedding-3-*} models, across three retrieval sizes ($k=3, 5, 10$). 
As shown in \autoref{table:rag}, providing higher-quality evidence, indicated by increased retrieval recall, generally enhances LLM performance in the RAG setting. 
Notably, the BM25 retriever achieves performance levels comparable to OpenAI’s embedding model, whereas the Contriever model significantly underperforms relative to both BM25 and OpenAI’s embedding models. 
Interestingly, despite BM25’s lower retrieval recall compared to OpenAI’s embedding model, the LLM generally achieves higher overall performance with BM25. 
This result highlights the robustness of effectiveness of term-based retrieval methods.

\subsection{Chain-of-Thought Analysis}
To better understand the effectiveness of CoT prompting methods for our tasks, we select several proprietary and open-source LLMs for experiments.  
In the w/o CoT setting, we instruct the LLMs to directly output the entailment label of the claim using the provided document context (The prompt used is provided in \autoref{fig:do_prompt} in appendix).  
As illustrated in \autoref{tab:cot_analysis}, both LLMs' performance degrades in the w/o CoT setting.
These results highlight the importance of CoT reasoning in enhancing LLM performance for our task.



\begin{table}[!t]
\centering 
\small
\addtolength{\tabcolsep}{-0.2em}
\resizebox{\linewidth}{!}{%
\begin{tabular}{lrccccc}
\toprule
Setting & $n$ & Recall &  Llama &  Mistral &  Qwen &  GPT-4o \\
\midrule
Contriever &     3 &  11.64 &           60.1 &           60.1 &         59.9 &    58.1 \\
BM25 &     3 &  31.96 &           67.7 &           67.3 &         69.6 &    65.7 \\
OAI-Large &     3 &  34.92 &           66.0 &           66.9 &         65.9 &    64.3 \\
\midrule
Contriever &     5 &  17.99 &           61.9 &           62.6 &         63.4 &    60.4 \\
BM25 &     5 &  38.62 &           71.9 &           73.1 &         73.1 &    71.6 \\
OAI-Large &     5 &  41.79 &           70.6 &           69.4 &         71.6 &    70.9 \\
\midrule
Contriever &    10 &  28.21 &           67.6 &           67.6 &         70.1 &    65.6 \\
BM25 &    10 &  46.42 &           73.4 &           75.7 &         78.1 &    73.9 \\
OAI-Large &    10 &  51.65 &           75.0 &           74.8 &         75.7 &    75.3 \\
\midrule
Oracle &   --- &     -- &           81.6 &           83.7 &         83.6 &    83.0 \\
\bottomrule
\end{tabular}
}

\caption{Performance comparison of various LLMs across different RAG settings on the \testmini set. The models evaluated include Llama-3.1-70B, Mistral-Large, Qwen2.5-72B, and GPT-4o.}
\label{table:rag}
\end{table}
\subsection{Error Analysis of Reasoning Process}
The \bestmodel model achieves a top accuracy of \best under the RAG setting. To better understand the model's limitations, we perform a detailed error analysis with human evaluators. We randomly select 25 instances from each of the three subsets where the \bestmodel model fails to perform correctly. 
Our analysis has identified four primary categories of errors:
(1) \emph{Extraction error}: The model fails to correctly locate the relevant context; additionally, it is also likely to extract data incorrectly from the table even when it identifies the relevant table. Both situations result in inaccurate verification.
(2) \emph{Numerical reasoning error}: The model encounters difficulties with correct mathematical reasoning. 
(3) \emph{Domain knowledge deficiency}: The model lacks sufficient knowledge in finance-related areas, which hampers its ability to reason accurately.
(4) \emph{Computation error}: While the model's reasoning is correct, it makes computational mistakes during intermediate or final steps, resulting in incorrect claim verification.

While our analysis is restricted to small-scale human evaluation, we believe that future work could explore advanced automated evaluation methods~\cite{liu-etal-2023-g, zheng2023judging, kamoi2024evaluating} for detecting reasoning errors within the generated explanations.

\begin{table}[!t]
\centering
\small
\addtolength{\tabcolsep}{-0.2em}
\resizebox{\linewidth}{!}{%
\begin{tabular}{lcccc}
\toprule
\multirow{2}{*}{\textbf{Model}} & \multicolumn{2}{c}{\textbf{w/o CoT}} & \multicolumn{2}{c}{\textbf{w/ CoT}}\\
\cmidrule(lr){2-3} \cmidrule(lr){4-5}
& LongC & RAG & LongC & RAG\\

\midrule
GPT-4o           & 78.1 \down{ 4.3} & 69.4 \down{ 5.9} & 82.4 & 75.3 \\
Qwen2.5-7B      & 51.1 \down{17.9} & 61.6 \down{10.8} & 69.0 & 72.4 \\
Ministral-8B     & 49.6 \down{ 3.1} & 59.4 \down{ 8.5} & 52.7 & 67.8 \\
Qwen2-7B         & 49.7 \down{ 3.9} & 62.1 \down{ 5.3} & 53.6 & 67.4 \\
Llama-3.1-8B     & 51.7 \down{15.2} & 61.6 \down{ 4.8} & 66.9 & 66.4 \\
Llama-3.2-3B     & 48.1 \down{ 6.0} & 50.6 \down{ 7.8} & 54.1 & 58.4 \\

\bottomrule
\end{tabular}
}

\caption{Comparison of LLM performance with and without CoT Prompting methods on the \testmini set.}
\label{tab:cot_analysis}
\end{table}
\section{Conclusion}
This paper presents \ours, a comprehensive benchmark designed to evaluate LLMs in claim verification over long and hybrid-content financial documents. Through extensive experiments involving \modelnum LLMs under long-context and RAG settings, we have demonstrated that even the top-performing models exhibit a significant performance gap compared to financial experts. Our detailed findings and insights reveal the strengths and limitations of current LLMs in this new task. We believe that \ours provides a valuable benchmark for future research on LLMs' ability to handle complex claim verification tasks within the expert domain.
\section*{Limitations}
First, our evaluation does not include recently released finance-specific LLMs~\cite{Wu2023BloombergGPTAL, yang2023fingpt, xie2023pixiu, xie2024openfinllm}, as these models are not yet compatible with the vLLM framework used for inference.
Additionally, due to computational resource constraints, we did not perform large-scale fine-tuning of LLMs on finance-domain data. However, we believe that training on such domain-specific data could significantly enhance LLM performance in \ours, particularly in terms of accuracy on the \knowset.
Moreover, we only conduct human error analysis on the generated reasoning process of models. We believe future work could explore the development of LLM-based automated evaluation systems~\cite{liu-etal-2023-g, zheng2023judging, kamoi2024evaluating} for automatically detecting reasoning errors within the generated explanation.

\section*{Acknowledgements}
We are grateful for the compute support provided by the Microsoft Research's AFMR program and Together AI\footnote{\url{https://www.together.ai/}}.

\bibliography{anthology,custom,llm}

\appendix

\clearpage
\section{Appendix}
\label{app:data}\label{app:data_example}\label{app:exp}
\label{app:error_analysis_examples}

\begin{figure}[h]
\begin{tcolorbox}[colback=black!3!white, colframe=black!70!white, title=An example within \ours \testmini set, fontupper=\footnotesize, fonttitle=\footnotesize]

\textbf{[Claim]} \\
The net interest expense in 2023, considering debt interest and capitalized interest, was \$183,479,000.
\newline
\newline
\textbf{[Supporting Evidence (Context Index)]}\\
107
\newline
\newline
\textbf{[Explanation of Reasoning Process]}\\
1. From table 107, the debt interest paid in 2023 is given as \$183,479,000 and the capitalized interest amount is \$-2,483,000).\\
2. We can calculate the net interest expense as 183479 - 2483 = 180996.\\
3. Therefore, the statement is refuted.

\end{tcolorbox}
\caption{An example within \ours \testmini set
}
\label{fig:detailed-example}
\end{figure}
\begin{figure}[h]
\begin{tcolorbox}[colback=black!3!white, colframe=black!70!white, title=Adopted Chain-of-Thought Prompt, fontupper=\footnotesize, fonttitle=\footnotesize]

\textbf{[System Input]} \\
As a financial expert, your task is to assess the truthfulness of the given claim by determining whether it is entailed or refuted based on the provided financial document. Follow these steps:\\
1. Carefully read the given context and the claim. \\
2. Analyze the document, focusing on the relevant financial data or facts that related to the claim.\\
3. Document each step of your reasoning process to ensure your assessment is clear and thorough.\\
4. Conclude your analysis with a final determination. In your last sentence, clearly state your conclusion in the following format: "Therefore, the claim is \{entailment\_label\}." Replace \{entailment\_label\} with either 'entailed' (if the claim is supported by the document) or 'refuted' (if the claim contradicts or partially contradicts the document).\\
\newline
\newline
\textbf{[User Input]}\\
Financial Report:\\
\textcolor{blue}{\{Financial Report\}} \\

Claim to verify:\\
\textcolor{blue}{\{Claim\}}
\newline\newline
Follow the instructions and think step by step to verify the claim.

\end{tcolorbox}
\caption{The Chain-of-Thought prompt used.
}
\label{fig:cot_prompt}
\end{figure}
\begin{figure}[h]
\resizebox{0.47\textwidth}{!}{
\begin{tcolorbox}[colback=black!3!white, colframe=black!70!white, title=Adopted Chain-of-Thought Prompt, fontupper=\footnotesize, fonttitle=\footnotesize]

\textbf{[System Input]} \\
As a financial expert, your task is to assess the truthfulness of the given statement by determining whether it is entailed or refuted based on the provided financial document. You should directly output the entailment label (`entailed' or `refuted') without any intermediate steps.\\
\newline
\newline
\textbf{[User Input]}\\
Financial Report:\\
\textcolor{blue}{\{Financial Report\}} \\

Claim to verify:\\
\textcolor{blue}{\{Claim\}}
\newline\newline
Directly output the entailment label (`entailed' or `refuted') of the claim.

\end{tcolorbox}
}
\caption{The \emph{Direct Output} prompt used.
}
\label{fig:do_prompt}
\end{figure}
\begin{table}[h]
\centering
\small
\resizebox{\linewidth}{!}{%
\begin{tabular}{lccc}
\toprule
\textbf{Annotation Quality}    & \textbf{\%S $\geq$ 4}\\
\midrule
\textbf{Claim} \\
\quad Fluency  & 92   \\
\quad Meaningfulness & 90\\
\quad Alignment with real-world scenarios & 94\\
\noalign{\vskip 0.5ex}\hdashline\noalign{\vskip 0.5ex}
\textbf{Evidence} \\
\quad Relevancy & 89 \\
\quad Completeness &85 \\
\noalign{\vskip 0.5ex}\hdashline\noalign{\vskip 0.5ex}
\textbf{Reasoning-process Explanation} \\
\quad Fluency &95 \\
\quad Correctness &92 \\
\quad Comprehensiveness &90 \\
\noalign{\vskip 0.5ex}\hdashline\noalign{\vskip 0.5ex}
\textbf{Entailment Label}\\
\quad Correctness &94\\
\bottomrule
\end{tabular}
}

\caption{Human evaluation over 100 samples from the \ours \testmini set. Two internal evaluators were asked to rate the samples on a scale of 1 to 5 individually. We report percent of samples that have an average score $\geq$ 4 to indicate the annotation quality of \ours.}
\label{tab:annotation_aggrement}
\end{table}
\begin{table*}[h]
\centering
\small
\resizebox{\textwidth}{!}{%
\begin{tabular}{clllc}
\toprule
\textbf{ID} & \textbf{Finance Industry Experience} & \textbf{English Proficiency} & \textbf{Annotation Sets} & \textbf{Evaluator?}\\
\midrule
1 & 1 working and 1 internship at US & Native speaker & \knowset &  \cmark \\

2 & >= 2 internship at US & > 15 years & \mathset & \cmark\\

3 & 1 working at Singapore and 2 internship at US & Native speaker & \knowset &  \cmark \\

4 & 2 working and >= 1 internship at US & Native speaker & \knowset & \xmark \\

5 & 1 internship at US, 2 internship at China & 10 years & \ieset & \xmark \\

6 & 1 internships at HK, China & 15 years & \ieset, \mathset & \cmark \\

7 & 1 internships at China & 10 years & \ieset, \mathset & \xmark\\
\bottomrule
\end{tabular}
}
\caption{Details of annotators involved in dataset construction. \ours is annotated by financial professionals with extensive experience in comprehending financial documents, ensuring it accurately reflects the real-world challenges in the financial domain.}
\label{tab:candidate_profiles}
\end{table*}

\begin{table*}[h]
\centering
\small
\begin{tabular}{p{0.28\textwidth}l@{\hspace{0.2cm}}l@{\hspace{0.2cm}}l@{\hspace{0.2cm}}p{0.3\textwidth}}
\toprule
\textbf{Model} & \textbf{Organization} & \textbf{Release Time} & \textbf{Max Length} & \textbf{Source} \\
\midrule
GPT-4o \cite{OpenAI2023GPT4TR}  & OpenAI   &   2024-03    & 128k & gpt-4o-2024-08-06 \\
\midrule
Gemini-1.5-Pro~\cite{geminiteam2024gemini}  & Google   &   2024-02    & 128k &  \\
\midrule
Claude-3.5-Sonnet~\cite{claude3}  & Anthropic   &  2024-03    & 200k & claude-3-5-sonnet-20241022 \\
\midrule
Llama-3.1~\cite{touvron2023llama} & Meta   &   2023-06    & 128k & meta-llama/Llama-3.1-*B-Instruct \\
Llama-3.2~\cite{meta2024introducing} & Meta   &   2024-09    & 128k & meta-llama/Llama-3.2-3B-Instruct \\
\midrule
Qwen-2~\cite{qwen2} & Qwen   &   2024-06    & 128k & Qwen/Qwen2-*-Instruct  \\
Qwen-2.5~\cite{qwen2.5} & Qwen   &   2024-09    & 128k & Qwen/Qwen2.5-*B-Instruct  \\
\midrule
Mistral~\cite{jiang2023mistral, jiang2024mixtral} & Mistral AI   &   2024-05    & 32k & mistralai/Mistral-7B-Instruct-v0.3  \\
Mathstral & Mistral AI   &   2024-08    & 32k & mistralai/Mathstral-7B-v0.1  \\
Ministral & Mistral AI   &   2024-10    & 128k & mistralai/Ministral-8B-Instruct-2410  \\
\midrule
InternLM2.5~\cite{cai2024internlm2} & internlm  &   2024-08    & 200k & internlm/internlm2\_5-7b-chat \\
\midrule
DeepSeek-V2~\cite{deepseekai2024deepseekv2} & deepseek AI  &   2024-05    & 128k & deepseek-ai/DeepSeek-V2-Lite-Chat \\
\midrule
GLM~\cite{du-etal-2022-glm} & 
THUDM &   2024-06    & 128k & THUDM/glm-4-9b-chat \\

\bottomrule
\end{tabular}

\caption{Details of the organization, release time, maximum context length, and model source (\ie url for proprietary models and Huggingface model name for open-source models) for the LLMs evaluated in \ours.}
\label{tab:model}
\end{table*}

\end{document}